\documentclass[conference]{IEEEtran}
\IEEEoverridecommandlockouts

\usepackage{cite}
\usepackage{amsthm}
\usepackage{amsmath,amssymb,amsfonts}
\usepackage{algorithmic}
\usepackage{graphicx}
\usepackage{textcomp}
\usepackage{xcolor}
\usepackage{hyperref}
\def\BibTeX{{\rm B\kern-.05em{\sc i\kern-.025em b}\kern-.08em
    T\kern-.1667em\lower.7ex\hbox{E}\kern-.125emX}}

\theoremstyle{definition}

\theoremstyle{remark}

\usepackage{enumitem}

\begin{document}

\title{Process Discovery Using Graph Neural Networks}

\author{\IEEEauthorblockN{Dominique Sommers, Vlado Menkovski, Dirk Fahland}
\IEEEauthorblockA{Eindhoven University of Technology, Mathematics and Computer Science, Eindhoven, the Netherlands\\
\{\href{mailto:d.sommers@tue.nl}{d.sommers}, \href{mailto:v.menkovski@tue.nl}{v.menkovski}, \href{mailto:d.fahland@tue.nl}{d.fahland}\}@tue.nl}}
\maketitle

\begin{abstract}
Automatically discovering a process model from an event log is the prime problem in process mining. This task is so far approached as an \emph{unsupervised} learning problem through graph synthesis algorithms. Algorithmic design decisions and heuristics allow for efficiently finding models in a reduced search space. However, design decisions and heuristics are derived from assumptions about how a given behavioral description\,---\,an event log\,---\,translates into a process model and were not learned from actual models  which introduce biases in the solutions. In this paper, we explore the problem of \emph{supervised learning} of a process discovery technique $d$. We introduce a technique for training an ML-based model $d$ using graph convolutional neural networks; $d$  translates a given input event log into a sound Petri net. We show that training $d$ on synthetically generated pairs of input logs and output models allows $d$ to translate previously unseen synthetic and several real-life event logs into sound, arbitrarily structured models of comparable accuracy and simplicity as existing state of the art techniques. We analyze the limitations of the proposed technique and outline alleys for future work.
\end{abstract}


\section{Introduction}\label{sec:introduction}

Automated process discovery (APD) is the problem of discovering a process model $M$ from an event log $L$~\cite{DBLP:reference/bdt/Leemans19}. State-of-the-art techniques approach APD as an \emph{unsupervised} learning problem trying to achieve pareto-optimality of $M$ regarding \emph{fitness} and \emph{precision} wrt. $L$, \emph{generalization} wrt. future traces not seen in $L$ yet, and structural \emph{simplicity} of $M$~\cite{DBLP:journals/ijcis/BuijsDA14}, and further to ensure \emph{soundness} of $M$~\cite{DBLP:journals/cj/VerbeekBA01}.  APD of flow-based models, such as Petri nets, is primarily approached algorithmically by synthesizing a graph from behavioral abstractions of $L$~\cite{DBLP:conf/bpm/LeemansFA13,DBLP:journals/dss/BrouckeW17,Augusto2018AutomatedApproach}, as optimization problem over linear~\cite{DBLP:journals/fuin/derWerfDHS09} or logical constraints~\cite{DBLP:journals/sosym/SoleC18}, or genetic algorithms searching for optima in the space of models~\cite{DBLP:journals/ijcis/BuijsDA14}.

Reviews and benchmarks observe that, despite impressive progress, no \emph{unsupervised APD technique} consistently returns fitting, precise, simple, \emph{and} sound model on all problem instances in feasible time~\cite{DBLP:journals/is/WeerdtBVB12,Augusto2019AutomatedBenchmark}. Specifically, each technique is based on different algorithmic design decisions and uses different heuristics for efficiently finding models in the available search space, resulting in an inherent bias favoring some quality criteria over others that cannot be overcome~\cite{Augusto2019AutomatedBenchmark}. These design decisions and heuristics are derived from assumptions about how a given behavioral description\,---\,an event log\,---\,translates into a process model in an ad-hoc manner and not derived systematically leading to low generalization~\cite{DBLP:conf/caise/WerfPWBR21} as we discuss in Sect.~\ref{sec:related_work}.

In contrast, human modelers train their modeling skills in a \emph{supervised} fashion by learning which model structures are adequate solutions for which behavior, and then apply these skills to build a solution piece-wise along the information provided~\cite{DBLP:conf/er/PinggeraZWFWMR10,PinggeraZWFWMR2011_er-bpm_ppm,Pinggera2013StylesIB}. In this paper, we study whether it is possible to design a process discovery technique that more directly emulates a human modeler, i.e. (1) \emph{learn} how to construct models from examples of event logs and corresponding models, and then (2) \emph{transfer} this learned knowledge to construct process models to event logs not seen previously.

We formulate the \emph{supervised process discovery problem}. We want to develop a technique $t$ that can train for given pairs $\langle L_i,M_i \rangle_{i=1}^k$ of (synthetically generated) event logs and corresponding process models a function $d = t(\langle L_i,M_i \rangle_{i=1}^k)$ so that $d$ can translate an (unseen, real-life) event log $L$ into a sound model $M = d(L)$ that has high accuracy wrt. $L$ and is structurally simple.

In the following, we propose a first solution to this problem for Petri nets as target modeling language where $d = (f,N_1,\ldots,N_k)$ is an algorithm $f$ constructing and updating a graph $G$ using graph convolutional neural networks $N_1,\ldots,N_k$. $f$ first encodes the general translation task from given log $L$ into the space of possible models as a graph $G$. $G$ encodes the input log $L$ with edges to a template Petri net $M$ having one transition $t_a$ per activity label $a$ in $L$ and candidate places and arcs between these transitions. $f$ then uses $N_1,\ldots,N_k$ to select which candidate places shall remain in the model based on the information in $L$. The $N_i$ update state vectors at event and transition nodes and at edges between nodes of $G$ to propagate behavioral information from the event log to the transitions. The technique $t$ trains $N_1,\ldots,N_k$ on $\langle L_i,M_i \rangle_{i=1}^k$ by following an iterative approach based on the process of process modeling observed in human modelers~\cite{PinggeraZWFWMR2011_er-bpm_ppm} to learn which places of target model $M_i$ shall remain given the structure of the input event log $L_i$. Through training, the $N_1,\ldots,N_k$ learn how to update the weights in $G$ to select places from the available candidates; $f$ uses beam search~\cite{graves2012sequence} and S-coverability checks~\cite{DBLP:journals/cj/VerbeekBA01} to prune the search space.

We implemented the above technique using DGL\footnote{\url{https://docs.dgl.ai/}} with PyTorch and demonstrate feasibility reaching state-of-the-art performance on a variety of problem instances. We trained $d$ on a synthetically generated training set of 2000 block-structured process models and corresponding event logs of varying size and complexity. We show that it is able to learn to rediscover the process models from the class of problems with same representational bias as it has been trained on, achieving high accuracy both wrt. the log and wrt. the data generating process that is comparable to the state-of-the-art techniques. With further evaluation on real-life processes, we show that our technique can effectively generalize to problem instances outside the representational bias of the training data, however for some instances the resulting model may only be easy sound (contain dead parts but no deadlocks). We specifically explore the reasons for the techniques current shortcomings and outline future avenues of research.

\section{Related work}\label{sec:related_work}
We discuss literature on quality measures, biases and design decisions in APD, modeling as a human task, and graph neural networks.


The quality of a process model $M$ to an event log $L$ is assessed along 5 criteria. \emph{Fitness} is the share of process executions (\emph{traces}) of $L$ that is described or accepted by $M$; \emph{precision} is the share of process executions described by $M$ that is not in $L$. \emph{Alignment-based fitness} and estimating precision via \emph{escaping edges} (only the first steps of $M$ not in $L$)~\cite{DBLP:books/sp/CarmonaDSW18} are most widely used~\cite{Augusto2019AutomatedBenchmark}. The monotone precision/recall measures~\cite{DBLP:journals/tosem/PolyvyanyySWCM20} rank models more consistently but are slower to compute. \emph{Generalization} is the likelihood that $M$ accepts another unseen trace from the process that generated $L$ and can only be estimated~\cite{DBLP:books/sp/CarmonaDSW18}. \emph{Simplicity} states how clearly $M$ describes the logic or cause-effect relations of the process that generated $L$ and is estimated through graph size and complexity~\cite{Augusto2019AutomatedBenchmark}. $M$ is \emph{sound} if every partial execution from the initial state can be extended to a terminating execution in a designated final state~\cite{DBLP:journals/cj/VerbeekBA01}. A sound workflow net $N$ can be decomposed into so-called \emph{S-components}, conversely if $N$ cannot be decomposed into \emph{S-components}, i.e., is \emph{not S-coverable} it is not sound. We exploit this property to exclude non-sound solutions.

The fundamental challenge in APD is that the target search space of models is too large to exhaust~\cite{DBLP:journals/fuin/derWerfDHS09,DBLP:journals/ijcis/BuijsDA14}. Genetic algorithms~\cite{DBLP:journals/ijcis/BuijsDA14} effectively explore the search space but take too long to find a satisfactory model for practical applications~\cite{Augusto2019AutomatedBenchmark}. Time-efficient algorithmic solutions to APD synthesize a graph from behavioral abstractions of the log~\cite{DBLP:conf/bpm/LeemansFA13,DBLP:journals/dss/BrouckeW17,Augusto2018AutomatedApproach}. Thereby design decisions and heuristics bias the algorithm regarding fitness, precision, simplicity, and soundness. Enforcing a specific \emph{representational bias}~\cite{DBLP:conf/bpm/AalstAM11} in the problem formulation, e.g., limiting the search space to block-structured models~\cite{DBLP:conf/bpm/LeemansFA13,DBLP:journals/ijcis/BuijsDA14}, ensures simplicity and soundness, although sound models with high fitness and precision lie outside the chosen representational bias~\cite{Augusto2019AutomatedBenchmark}. Algorithmic design decisions can favor or even guarantee solutions of a specific quality criterion at the cost of loss in another criterion, e.g., ensuring fitness~\cite{DBLP:conf/bpm/LeemansFA13,DBLP:journals/fuin/derWerfDHS09,DBLP:journals/sosym/SoleC18} lowers precision~\cite{DBLP:journals/sosym/SoleC18,Augusto2019AutomatedBenchmark}. Heuristics-based filtering and pattern detection on behavioral abstractions of event data results in models of high fitness and precision~\cite{DBLP:journals/dss/BrouckeW17,Augusto2018AutomatedApproach} which in turn may be unsound and have high complexity~\cite{Augusto2019AutomatedBenchmark}; the heuristics may not generalize to new data~\cite{DBLP:conf/otm/LeemansTH18} or larger samples~\cite{DBLP:conf/caise/WerfPWBR21}. Techniques relying on behavioral abstractions of event logs~\cite{DBLP:conf/bpm/LeemansFA13,DBLP:journals/dss/BrouckeW17,Augusto2018AutomatedApproach,DBLP:journals/tkde/AalstWM04} fail when the log contains behaviors not preserved by the abstraction~\cite{DBLP:journals/kais/LeemansF20,DBLP:journals/tkde/AalstWM04}. Techniques avoiding behavioral abstractions solve an optimization problem over linear~\cite{DBLP:journals/fuin/derWerfDHS09,DBLP:conf/icpm/Bergenthum19} or logical constraints~\cite{DBLP:journals/sosym/SoleC18} over the event log that ensures fitness at the cost of precision and soundness~\cite{DBLP:journals/sosym/SoleC18,Augusto2019AutomatedBenchmark} or prohibitively high running times~\cite{DBLP:conf/icpm/Bergenthum19}.

\label{sec:modeling}The cognitive process of humans creating models has been studied empirically. Humans create models by iterating three phases: \emph{comprehend} (a chunk of the information about the process), \emph{model} (by adding or removing formal model structures related to the comprehended chunk), and \emph{reconcile} (by reorganizing model layout to better comprehend the created model structures)~\cite{PinggeraZWFWMR2011_er-bpm_ppm}. Empirical evaluations have shown that breadth-first modeling strategies along the structure of the target model lead to highest precision and recall of the input information~\cite{DBLP:conf/er/PinggeraZWFWMR10}.

Machine learning methods have not been exploited for APD as of yet. However, neural networks have been developed for various graph problems~\cite{kipf2016semi}. Such networks learn representations of graphs, node and edges based on an information propagation process to perform tasks like node classification or link prediction. Generative graph neural networks are developed that learn and represent conditional structures on graphs~\cite{li2018learning}. The attention mechanism have proven useful in generative tasks that can redistribute the weights of the different inputs~\cite{vaswani2017attention} and has been exploited in graph neural networks as well~\cite{velivckovic2017graph}.

\section{Defining the learning task}
\begin{figure}[t]
\centerline{\includegraphics[width=0.9\linewidth]{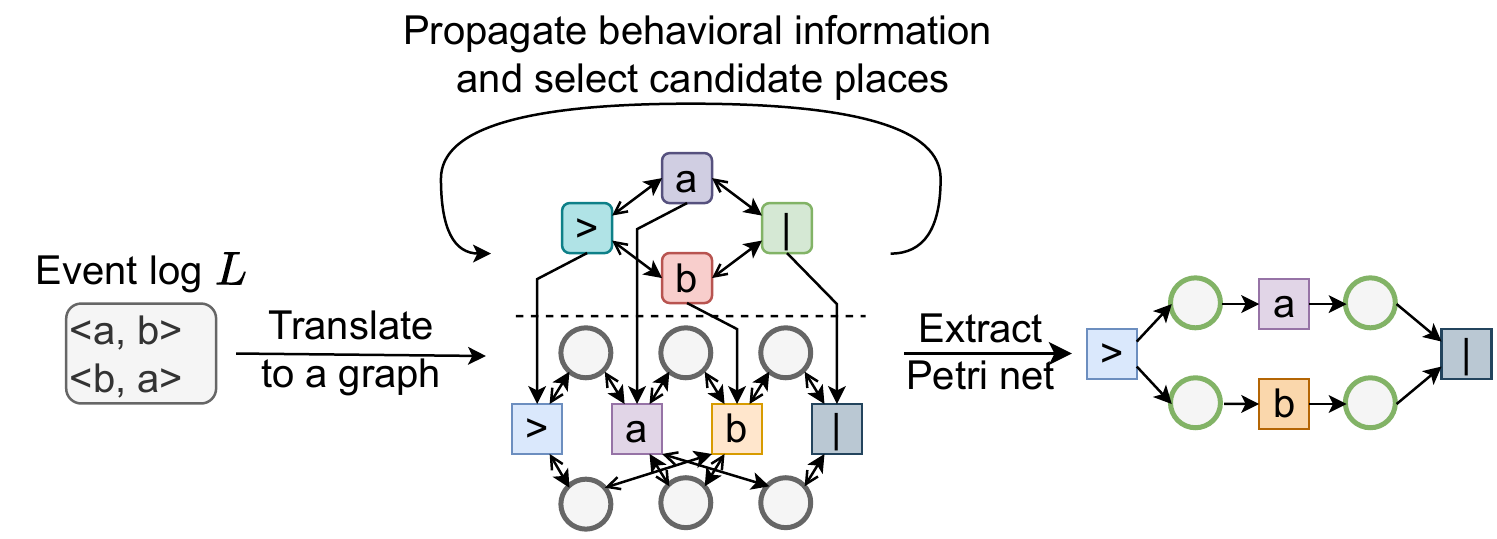}}
\caption{High level overview of the approach.}
\label{fig:overview_highlevel}
\end{figure}
The fundamental task of process discovery is to translate an event log $L$ into a Petri net $N$ (see Fig. \ref{fig:overview_highlevel}). In this section, we show how to formally encode this discovery problem in a graph $G$ (Sect. \ref{sec:data_prep}) which contains both the event log $L$ and a candidate Petri net $N$ with candidate places $P$. We then define the learning task for solving the discovery problem (Sect \ref{sec:learning_task}) which is to select the candidate places from $P$ that most likely explain the behavioral information in $L$; the sub-graph with the selected candidate places is the resulting model $N$. In Sect. \ref{sec:approach}, we explain how to train a function $d$ to learn how to solve this discovery problem in an iterative manner by propagating behavioral information from $L$ to model structures in the candidate Petri net $N$ and select candidate places one-by-one. Fig. \ref{fig:overview_highlevel} summarizes this approach.

\subsection{Preliminaries}
We write $A$ for the (finite) set of \emph{activity names} extended with $\mathord{>}\in A$ and $\mathord{|} \in A$ for artificial start and end. A \emph{log} $L$ (over $A$) is a finite multiset of traces where a \emph{trace} is a finite sequence $\sigma = \langle >, a_1,\ldots,a_n,| \rangle \in A^*$. Each occurrence of an activity $a_i \in \sigma$ is called an \emph{event}.

We recall notations for Petri nets with $\tau$-transitions and refer to~\cite{DBLP:journals/cj/VerbeekBA01} for definitions. A Petri net $N = (P,T_A \cup T_\tau,F)$ has places $P$, transitions $T = T_A \cup T_\tau$ where $T_A$ and $T_\tau$ are \emph{visible} and \emph{invisible} transitions (i.e., the $T_\tau$ do not occur in firing sequences of $N$), respectively, and arcs $F$. In a \emph{workflow net} $N$, every node of $N$ is on a path (along $F$) from the unique source place $i \in P$ (no incoming arcs) to the unique sink place $o$ (no outgoing arcs).

\subsection{Encoding the discovery problem in a graph}\label{sec:data_prep}
\begin{figure}[t]
\hspace*{-5mm}
\centerline{\includegraphics[width=1.1\linewidth]{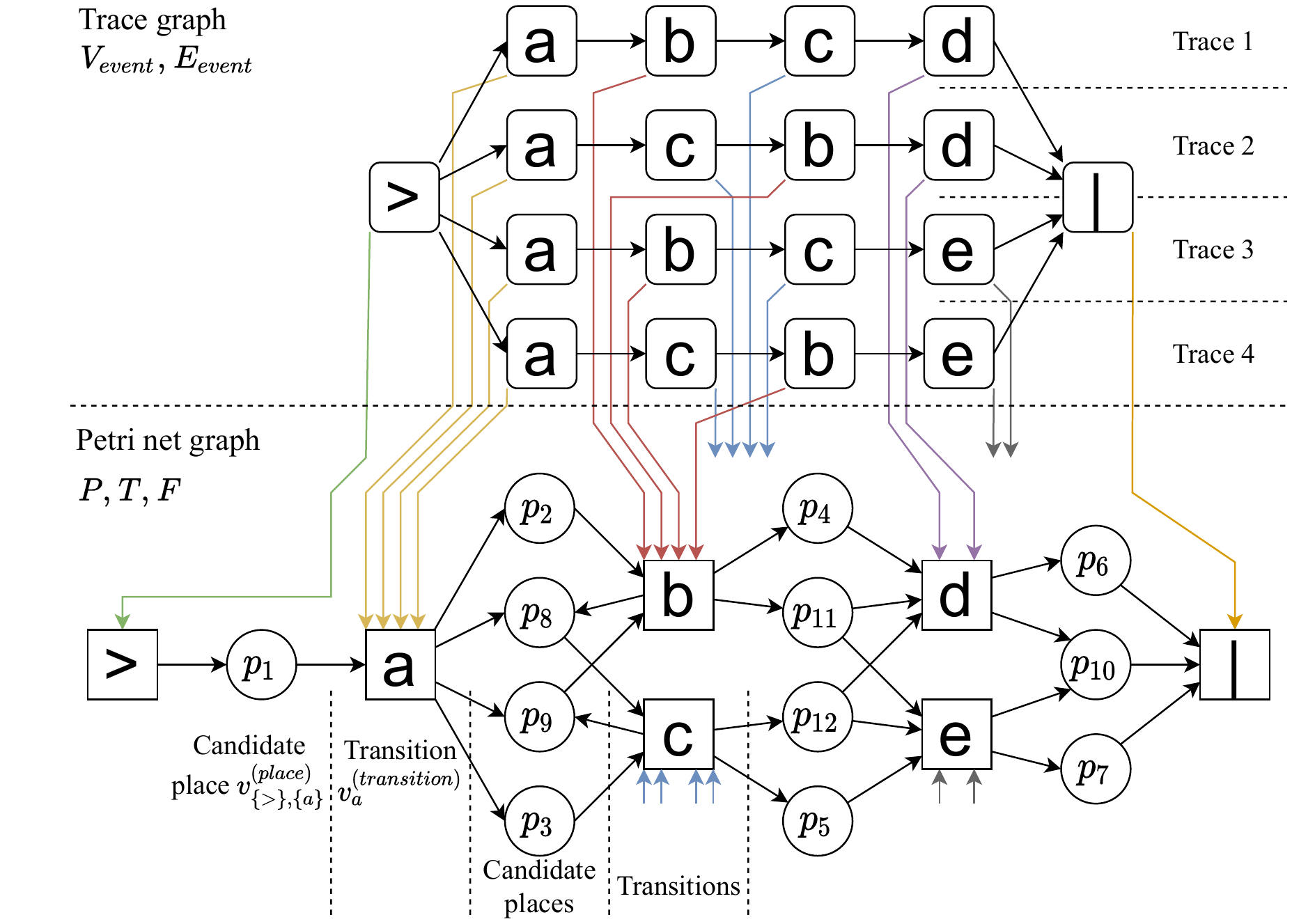}}
\caption{Encoding of process discovery problem.}
\label{fig:data_preparation}
\end{figure}

We encode the problem of translating a given input event log $L$ into a Petri net $N$ as a graph $G$ of 3 parts as illustrated in Fig.~\ref{fig:data_preparation}. (1) We encode $L$ as a \emph{trace graph}. (2) We encode the solution space of all possible models for $N$ as a \emph{candidate Petri net} (over-approximating the required places and arcs). (3) \emph{Links} from event nodes in the trace graph to transitions in the candidate Petri net encode which transition shall describe which event.


To encode $L$, the \emph{trace graph} $(V_{event},E_{event})$ contains one start and one end event node $v_{\mathord{>}}, v_{\mathord{|}} \in V_{event}$, and for each trace $\sigma = \langle \mathord{>},a_1,\ldots,a_n,| \rangle \in L$ the chain of unique event nodes $v_{a_i}^{\sigma,i}, 1 \leq i \leq n$ with directed edges $(v_{a_i}^{\sigma,i}, v_{a_{i+1}}^{\sigma,i+1})$, $(v_{a_{i+1}}^{\sigma,i+1},v_{a_i}^{\sigma,i}) \in E_{event}, 1 \leq i \leq n$ and directed edges between $v_{\mathord{>}}$ and $v_{a_1}^{\sigma,1}$ and between $v_{a_n}^{\sigma,n}$ and $v_{\mathord{|}}$.

The \emph{candidate Petri net graph} $(P,T,F)$ defines a superset of the candidate places $P$ and arcs $F$ needed to describe the behavior in $L$; within this superset we search for the target model $N$. $T$ contains one transition $t_a \in T$ for each activity $a \in A$ in $L$ ($\mathord{>},\mathord{|} \in A$). Similar to the $\alpha$-algorithm~\cite{DBLP:journals/tkde/AalstWM04}, we fully characterize each candidate place $p$ and its incoming and outgoing arcs by a pair $(X,Y), X,Y \subseteq T$ of input and output transitions. The pair $(X,Y)$ defines place $p_{X,Y}$ with arcs $(t,p_{X,Y}),t\in X, (p_{X,Y},t),t\in Y$. Within the set of all possible candidate places (and arcs) $P = \{ (X,Y) \mid X,Y \subseteq T \}$ we search the Petri net $(P',T,F')$, with places $P' \subseteq P$ and corresponding arcs $F' \subseteq F$ that best describe $L$. We can use the $\alpha$ relations to exclude from $P$ candidates incompatible with $L$ and thus reduce the search space; see App.\ref{app:search_space_reduction}. 

\emph{Links} are edges from event to transition nodes. For each event node $v_{a_i}^{\sigma,i}$ we add a directed edge $(v_{a_i}^{\sigma,i},t_{a_i})$ to $t_{a_i}$. Thus, the graph structure encodes which transition nodes shall model which events. 

\subsection{Defining the learning task}\label{sec:learning_task}
Graph $G$ essentially defines the learning task of identifying only the subset of $P$ that restrict the transitions $T$ to the behavioral patterns present in $L$, e.g., choices, parallelism.


Where the $\alpha$-algorithm constructs the subset of $P$ from $L$, we train a function $d$ on examples $\langle L_i,N_i \rangle$ to (learn to) estimate the likelihood that a specific place $v_i \in P$ is best suited to be included in $N$ given that places $v_1,...,v_{i-1}$ were chosen before, i.e., $p(v_i|v_1,...,v_{i-1})$. However, we have to make sure that the entire set $P'$ describes the most likely solution for a specific $L$, i.e. the joint probability $p(P',\pi|L) = \prod_{i=1}^{\left| P' \right|} p(v_i|v_1,\dots,v_{i-1}, L)$ with ordering $\pi$. Note that the actual ordering of $P'$ is irrelevant in a graph and therefore ideally $p(P',\pi|L)$ is independent of $\pi$. \emph{Marginal joint probability} models this probability over all permutations $\mathcal{P}(P')$ of $P'$: $p(P'|L)=\sum_{\pi \in \mathcal{P}(P')} p(P',\pi|L)$. $p(P'|L)$ can be used to find $\arg\max P'$ for event log $L$ by sampling those resulting in the highest probability.

\section{Approach}\label{sec:approach}

Our approach for defining $d$ consists of a set of (graph) neural networks, each taking as input (a part of) $G$ and emulates a specific step in sequentially creating a model from an event log by propagating behavioral information and selecting candidate places (see Fig. \ref{fig:sequential_model_overview}). These networks together model the joint probability $p(P'|L)$ of selecting only the subset $P' \subseteq P$ needed to describe the behavioral relations between transition nodes available from an event log $L$.

We do not directly translate from $L$ to $p(P'|L)$ but introduce a latent parameter space to encode and propagate behavioral relations from $L$ to the nodes in $N$. In Sect. \ref{sec:seq_pm_gen}, we introduce a feature vector $h_i$ for each node $v_i \in V$ of $G$; $h_i$ allows to encode how $v_i$ is related to each activity $a \in A$ recorded in $L$. In \ref{sec:seq_pm_gen}, we also explain how each of the (graph) neural networks either updates $h_i$ to propagate information in $G$ (comprehension/reconciliation) or uses $h_i$ to learn how to maximize $p(P'|L)$ (i.e., select the most likely places $P'$). Sect \ref{sec:training} discusses how to train the NNs for maximizing $p(P'|L)$ together based on training data $\langle L_i,N_i \rangle$. Sect \ref{sec:extensions} introduces extensions to $f$ to ensure S-coverability and the possibility of incorporating invisible $\tau$-transitions.

\subsection{Sequential candidate selection}\label{sec:seq_pm_gen}
\begin{figure*}[htb]
\centerline{\includegraphics[width=1\linewidth]{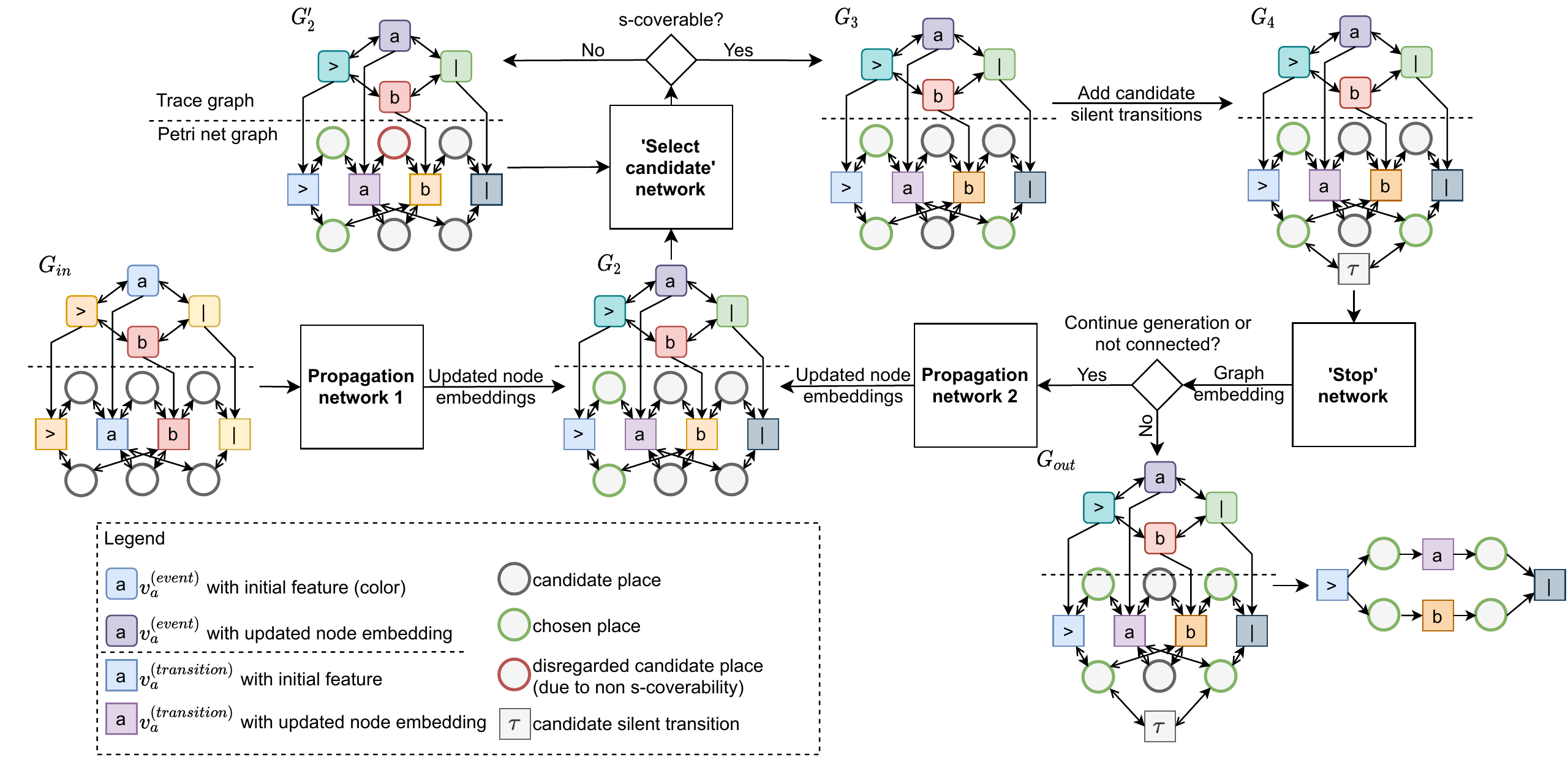}}
\caption{Detailed overview of the candidate selection.}
\label{fig:sequential_model_overview}
\end{figure*}

Function $d$ uses 2 decision making neural networks (NNs), ``Select candidate'' NN, and ``Stop'' NN, and 2 graph convolutional networks (GCNs), propagation networks PN1 and PN2 as follows. $d$ solves the task of sequentially identifying $P'$ in two steps: deciding which candidate place to select next (Sect. \ref{sec:approach_step_2}), and deciding when to stop selecting more places (Sect. \ref{sec:approach_step_3}). For both, a regular ``decision making'' NN is sufficient to model the probability distribution on nodes for selection; see Fig. \ref{fig:sequential_model_overview}.

However as is stated in above, these NNs have to make their decisions based on behavioral information from $L$ encoded in the trace graph in $G$. Graph convolutional neural networks (GCNs) can propagate and process information through $G$ to encode this behavioral information in a node's feature vector $h_i$ (also called \emph{node embedding}). Initially, each node $v_i$ is given a one-hot encoded feature vector $h_i^{(0)}$ of length $|A|$ denoting its activity label concatenated with its frequency in the event log, if available. $h_i^{(0)}$ for candidate places are zero vectors.

GCNs allow to update a node's feature vector $h_i$ based on the feature vectors of its (indirect) neighbors, taking the surrounding graph structure into account. This allows us to train a GCN PN1 to process and propagate the behavioral information from the event log nodes in $G$ to the place nodes in $G$ such that the NNs can make their decision for selection (Sect. \ref{sec:approach_step_1}).

As the method is sequential, more propagation is required by a second GCN PN2 through $G$ to aggregate the NN's past decisions as future decisions are conditionally dependent on them (Sect. \ref{sec:approach_step_4}). These four NNs now each have their own smaller learning task which align with the steps in the process of process modeling (see Section \ref{sec:modeling}).
Fig. \ref{fig:sequential_model_overview} shows how they are connected and modify $G$, which is similar to the graph generation process as proposed in~\cite{li2018learning}. The following sections go into detail of each NN's learning task.

\subsubsection{Propagation network 1}\label{sec:approach_step_1}
\begin{figure}[ht]
\centerline{\includegraphics[width=0.9\linewidth]{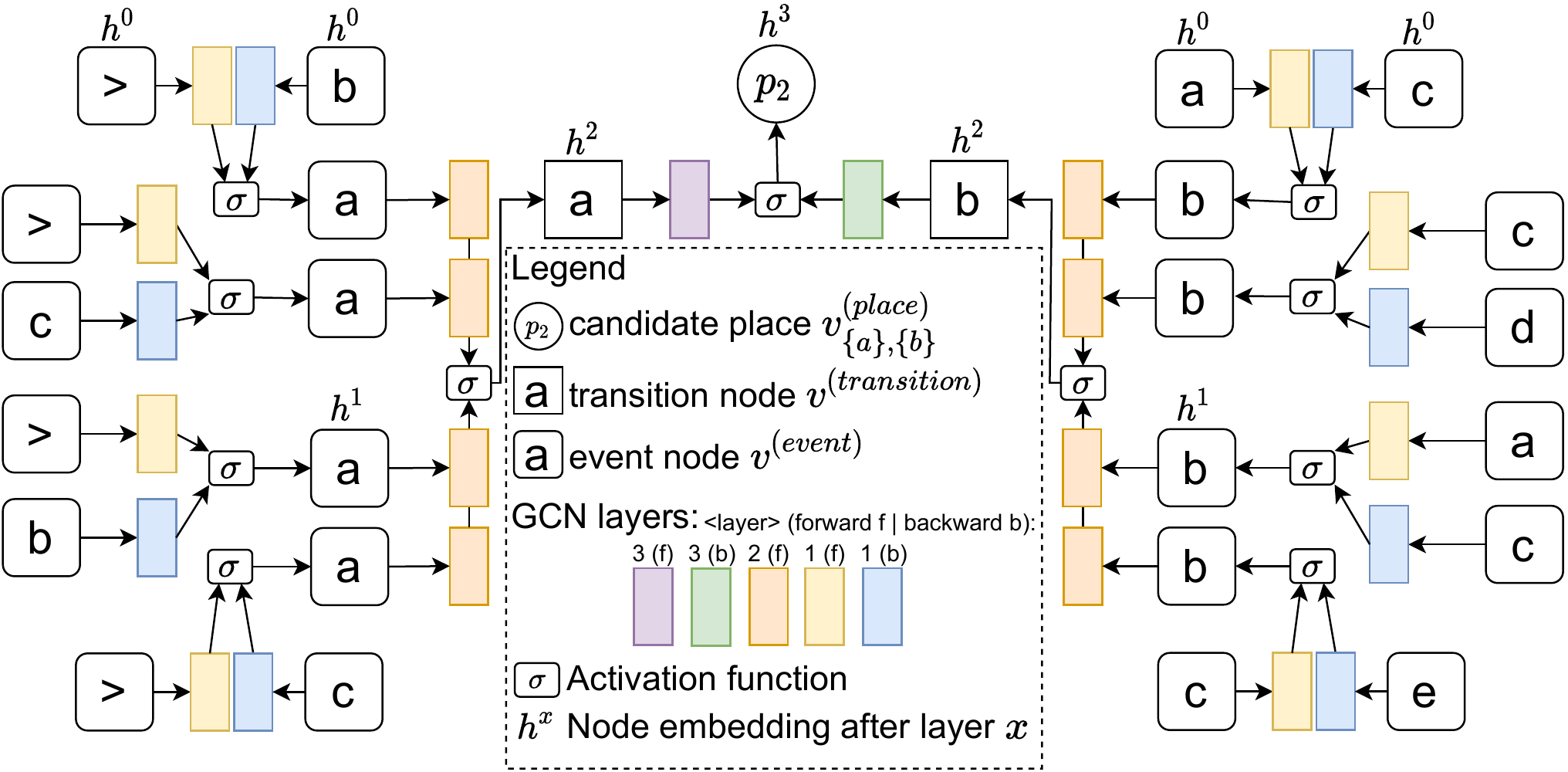}}
\caption{Partial illustration of information propagation through the graph for node $v_{\{a\},\{b\}}^{(place)}$ from Fig. \ref{fig:data_preparation}.}
\label{fig:gcn_place_propagation}
\end{figure}

We propagate behavioral information from  the trace graph to the places by a GCN with $K$-headed attention mechanism as described in \cite{velivckovic2017graph}. The number of layers $l$ in a GCN controls how many propagation steps are performed. Each place $p$ in $G$ has at least input transition $t_a$ and one output transition $t_b$; to at least propagate behavioral information from the predecessor and successor events of both $a$ and $b$, PN1 should contain at least three layers (as illustrated in Fig. \ref{fig:gcn_place_propagation}). The attention mechanism can be used to give unequal weights to nodes and is necessary because of the different node classes in $G$ which are not equally important for this initial step.

The result of this step is the same graph with updated node embeddings as computed by the internal weights of the GCNs as follows:

For the first ($l - 1$) layers of the network the nodes' embeddings are updated by:
\begin{equation}\label{eq:GCN_both_directions_first}
  h_i^{(l+1)} = \mathbin\Vert_{k=1}^K \left(  \sum_{j \in \mathcal{N}(i)} d_{ij} \begin{bmatrix} \alpha_{ij}^{(l)^k} h_j^{(l)} W^{(l)^k} \\ \alpha_{ij_r}^{(l)^k} h_j^{(l)} W^{(l)^k}_r  \end{bmatrix} \right)
\end{equation}

where $h_i^l$ denotes the embedding of node $v_i$ at layer $l$, $\mathcal{N}(i)$ denotes the \emph{outgoing} neighbors of node $v_i$ and $W^{(l)^k}$ is the weight matrix of the GCN at layer $l$ and attention head $k$. To also aggregate information from \emph{incoming} neighbors, all arcs in $G$, except for the links, are made bi-directional, where the vector $d_{ij}$ encodes the direction of the arc: $\left[1,0\right]$ if from $v_i$ to $v_j$, and $\left[0,1\right]$ if from $v_j$ to $v_i$. Furthermore, self-loops are added to retain a node's own information during aggregation. $W_r^{(l)^k}$ processes information on these \emph{reversed} arcs.

The update function of the last layer, having a single attention head, is:
\begin{equation}\label{eq:GCN_both_directions_last}
  h_i^{(l+1)} = \textsc{ReLU}\left(  \sum_{j \in \mathcal{N}(i)} d_{ij} 
  \begin{bmatrix} \alpha_{ij}^{(l)} h_j^{(l)} W^{(l)} \\ \alpha_{ij_r}^{(l)} h_j^{(l)} W^{(l)}_r \end{bmatrix} \right)
\end{equation}
with the \textsc{ReLU} activation function.

The nodes' embeddings $h_i$ are used by the GCN to encode the behavioral information and are only interpretable by the ``Select candidate'' and ``Stop'' NNs. to be later used for classification of the places. Such information is similar to how a human modeler would aggregate information from the event log and the place's structural properties to decide whether it fits the process.


\subsubsection{``Select candidate'' network}\label{sec:approach_step_2}
We now determine for each candidate place $v_i \in P \setminus P'$ that was not selected yet, the probability $p_i$ that $v_i$ should be added next, and pick $\arg\max_{v_i \in P \setminus P'} p_i$. $v_i$ is marked as selected by adding a feature value to the embedding.

To compute $p_i$, we define CN as a regular NN with input vector $h_i$ (as returned by the preceding step) and output a single value $p_i$. CN learns the weight matrix $W$ to map $h_i$ to $p_i$ as a single fully-connected layer.

\begin{equation}
    s_v = h_v W
\end{equation}
By normalizing all the candidates' scores using the \textsc{Softmax} function we get a probability $p_v$ for each node:
\begin{equation}
    p_v = \textsc{Softmax}(s_v) = \frac{\exp(s_v)}{\sum_{u \in (P\setminus P')} \exp(s_u) }
\end{equation}
with $\exp(x) = e^x$. 


\subsubsection{``Stop'' network}\label{sec:approach_step_3}
We define the ``Stop'' network SN for deciding when to stop adding candidate places. SN takes as input all Petri net node embeddings $h_i, v_i \in V \setminus V_{event}$ and has as output a probability $p_{add}$ that another place should be added; we then make the binary decision to stop adding by comparing to a chosen threshold. SN has two layers. Layer 1 aggregates $h_i, v_i \in V \setminus V_{event}$ into a graph embedding $h_G$ by the following equation

\begin{equation}
    h_G = \sum_{v \in V \setminus V_{event}} \textsc{Sigmoid}(h_v W_a) (h_v W_g)
\end{equation}
with learnable weight matrices $W_a$ and $W_g$. $\textsc{Sigmoid}(h_v W_a)$ serves as a ``gating function'', how much each node should contribute to $h_G$ by the weights in $W_a$~\cite{li2018learning}, where $\textsc{SIGMOID}(x)=\frac{1}{1 + e^{-x}}$ is used to map a value from $\mathbb{R}$ to a value between 0 and 1.

Layer 2 computes the probability $p_{add}$ to add more nodes from $h_G$ by:
\begin{equation}
    p_{add} = \textsc{Sigmoid}(h_G W_d)
\end{equation}
with a learnable weight matrix $W_d$. $(h_G W_d)$ is a score which is converted to a probability using the logistic sigmoid function. 

If the network's decision is to stop, we are done. Otherwise, we continue at Section \ref{sec:approach_step_4}.

\subsubsection{Propagation network 2}\label{sec:approach_step_4}
This step in the process is similar as the one in Section \ref{sec:approach_step_1} where information is propagated through the graph by a GCN with multi-headed attention mechanism and is used to process the previously made decision. It is necessary for the GCN to have at least two layers to ensure that information about the previous decision reaches other candidates. Recall that the decisions are encoded as features in the embeddings. By propagating this information two steps in the graph, $d$ can encode information whether other candidates fit the already selected candidates. The nodes' embeddings are updated in a similar fashion as is described in Section \ref{sec:approach_step_1}, after which we loop back to Section \ref{sec:approach_step_2} to make the next decision.

\subsection{Training and inference}\label{sec:training}
During training on instances $\langle L_j, N_j \rangle$ with known solutions $N_j = (P_j, T_j, F_j)$, the weight matrices in the NNs are optimized such that a specified loss function is minimized. The loss function to be minimized during training is as follows:
\begin{equation}\label{eq:negative_log_loss}
    l = - \sum_{i=1}^{\left| P'\right|} \log p(v_i|v_1,\dots,v_{i-1})
\end{equation}
with known places $P_j = P'\subseteq P$ to be selected and $p(v_i|v_1,\dots,v_{i-1})$ the probability of choosing candidate place $v_i$ after having selected candidate places $v_1$ to $v_{i-1}$. It is intractable to learn the marginal joint probability because many permutations may exist. A canonical ordering can give a lower bound. For learning, we choose $\pi$ based on breadth-first search on $N_j$; this order is close to canonical. With $\pi$ and Eq. \ref{eq:negative_log_loss}, the model learns the joint distribution $q = p(\mathcal{P}', \pi | \mathcal{L})$ of the data by maximizing the expected joint log-likelihood:
\begin{equation}
    \mathbb{E}_{p_{data}(\mathcal{P}',\pi|\mathcal{L})}\left[ \log q \right] = \mathbb{E}_{p_{data}(\mathcal{P}'|\mathcal{L})} \mathbb{E}_{p_{data}(\pi| \mathcal{P}')} \left[ \log q \right]
\end{equation}
with $\mathcal{P}' = \{N_j\}_{j=1}^k$ and $\mathcal{L}=\{L_j\}_{j=1}^k$ from a dataset $\langle L_j, N_j\rangle_{j=1}^k$. With a sufficiently large and varying dataset $\mathbb{E}_{p_{data}(\mathcal{P}',\pi|\mathcal{L})}\left[ \log q  \right]$ approaches $\mathbb{E}_{p(\mathcal{P}'|\mathcal{L})}\left[ \log p(\mathcal{P}' | \mathcal{L})  \right]$ which is the true probability distribution of the correct selection of candidates $P'$ from $P$ given an event log $L$, solving the learning task.

Teacher forcing~\cite{williams1989learning} is used in training to enforce the breadth-first ordering $\pi$ during the generation process and to correct the prediction after every candidate choice, which addresses slow convergence and instability.

During inference\,---\,the counterpart of training where $d$ is used on unseen data\,---\,teacher forcing is inaccessible and therefore the sampled candidates are selected by the NNs without correction. A suboptimal choice in the beginning of the generation process can lead to a low joint probability at the end, which is exactly what we want to maximize. To mitigate this, beam search is used; a heuristic search algorithm to find the highest joint probability. Using beam search with a beam width of $b$, $d$ selects the $b$ candidates with highest conditional probability at each step from $b$ unfinished runs, resulting in a set of $b^2$ (un)finished runs of which the $b$ runs with highest joint probabilities are taken for the next step. Afterwards, the $b$ Petri nets with highest joint probabilities are returned.

\subsection{Extensions}\label{sec:extensions}
As illustrated in Fig. \ref{fig:sequential_model_overview}, extensions on the technique as described above are made to ensure certain properties of the Petri net, also on real-life data.
Firstly, we do not add a candidate place $v_i$ chosen by SCN if the net $N$ with $v_i$ is not S-coverable (a polynomial check on the structure of $N$) as then the resulting net would not be sound \cite{DBLP:journals/cj/VerbeekBA01}; in this case, we exclude $v_i$ and let SCN return the next most probable candidate (see Fig. 3). This ensures reachability of the final marking but cannot avoid dead transitions.
To enforce proper soundness, a global check for a workflow net structure could be introduced.

Secondly, SN (Sect. \ref{sec:approach_step_3}) may decide to stop while some transitions do not have an incoming or outgoing place (not a workflow net); we check for non-connectedness of $N$ and override the stop decision in this case.

Invisible $\tau$-transitions are needed to model silent skips in real-life event logs while they are not recorded in $L$. We add $\tau$-transitions $T_\tau$ to the candidate nodes (previously $P$) for the SCN to select. In principle, the set of all possible candidate $\tau$-transitions is $T_{\tau} = \{(X,Y)|X,Y \subseteq P\}$, however, for feasibility, we only add candidate $\tau$-transitions between already selected candidate places: $T_{\tau} = \{(X,Y)|X,Y \subseteq P'\}$. Candidate $\tau$-transitions are added to $G$ after selecting a new place and its feature vector is defined as the sum of the feature vectors of its neighboring nodes. 

\section{Evaluation}\label{sec:evaluation}
We assess the feasibility of our approach for solving the supervised process discovery problem, i.e., training $d$ on $\langle L_i,N_i \rangle$ synthetic process instances to apply $d$ for discovering models on unseen synthetic and real-life data.
\subsection{Experimental setup}
We evaluate wrt. 3 objectives: (O1) Is the loss function of Eq. \ref{eq:negative_log_loss} able to optimize the NNs in $d$ to achieve high precision/recall in selecting the ground truth places on synthetic test data? (O2) Does the loss function also optimize fitness and precision of the discovered model wrt. the input model on synthetic test data compared to state-of-the-art methods? (O3) How does the quality of models compare to the other methods on real-life data, i.e., does the ML program generalize beyond the task it was trained on?
For training and testing, we used the PTAndLogGenerator plugin in ProM to generate $\langle L_i, N_i\rangle_{i=1}^{2663}$ problem instances: we generated 2663 random process trees, converted to block-structured Petri net $N_i$, and generated corresponding event log $L_i$ by simulation with the following parameters:

\begin{itemize}[label=\raisebox{0.25ex}{\tiny$\bullet$}]
    \item number of nodes: mode=8, min=4, max=15;
    \item structure probabilities: sequence=0.4, choice=0.32, parallel=0.2, loop=0.08, or=0.0;
    \item number of silent transitions: 4;
    \item number of simulated traces per Petri net: 1000.
\end{itemize}
The hyperparameters for our model $d$ are:
\begin{itemize}[label=\raisebox{0.25ex}{\tiny$\bullet$}]
    \item PN1: 4 layers with 20;32;64;32 neurons, output size 16;
    \item PN2: 2 layers with 17;32 neurons, output size 16;
    \item SCN: 1 layers with 17 neurons, output size 1;
    \item SN: 1 layers with 16 neurons, output size 1.
\end{itemize}
For (O1), $d$ is trained and tested on the dataset for 100 epochs, with a 75/25 (2000/663) train/test split, all while keeping track of the loss and percentages of true/false positive places. 

For (O2), inference is performed using the trained $d$ on the test dataset from (O1), with a beam width $b=10$. For the produced process models, we measure alignment-based fitness and precision \cite{DBLP:books/sp/CarmonaDSW18} to obtain F-scores and simplicity scores based on the inverse arc degree \cite{blum2015metrics}. We compared to Inductive Miner (IM) \cite{DBLP:conf/bpm/LeemansFA13}, Split Miner (SM) \cite{Augusto2018AutomatedApproach}, and Heuristics Miner (HM) \cite{weijters2006process}.

For (O3), inference is performed on eleven real-life datasets that are selected from the BPI challenge, with a beam width $b=50$. Because of the input size of 20 and the artificial start and end transition, only the 18 most frequent occurring activities are taken from each dataset. For both O2 and O3, we lowered beam width after each selected place. Furthermore, we only used a sample of size between 8 and 50 (depending on the size of $G$) from each log to keep the number of candidate places tractable. Note that conformance checking is always done on the complete event log.

All experiments are performed on an Intel i7-8705G CPU (3.10GHz) with 16 GB RAM an no GPU support; the source is available at \href{https://gitlab.com/dominiquesommers/apd-ml}{gitlab.com/dominiquesommers/apd-ml}.

\subsection{Results}
\emph{(O1)} the loss converges towards zero after the 100 epochs, but does not reach zero, which is expected since the ordering of candidates is not completely canonical (c.f Sect. \ref{sec:training}). Inference during training shows a similar trend for selecting true positive candidate places as the loss. The number of false positives keeps decreasing after the loss is converged. This is due to the use of teacher forcing causing a mismatch in training and inference. Note that the loss is directly related to the number of true positives, but not to the false positives as teacher forcing corrects $d$ after every selection during training.



\begin{figure}[t]
\centerline{\includegraphics[width=1\linewidth]{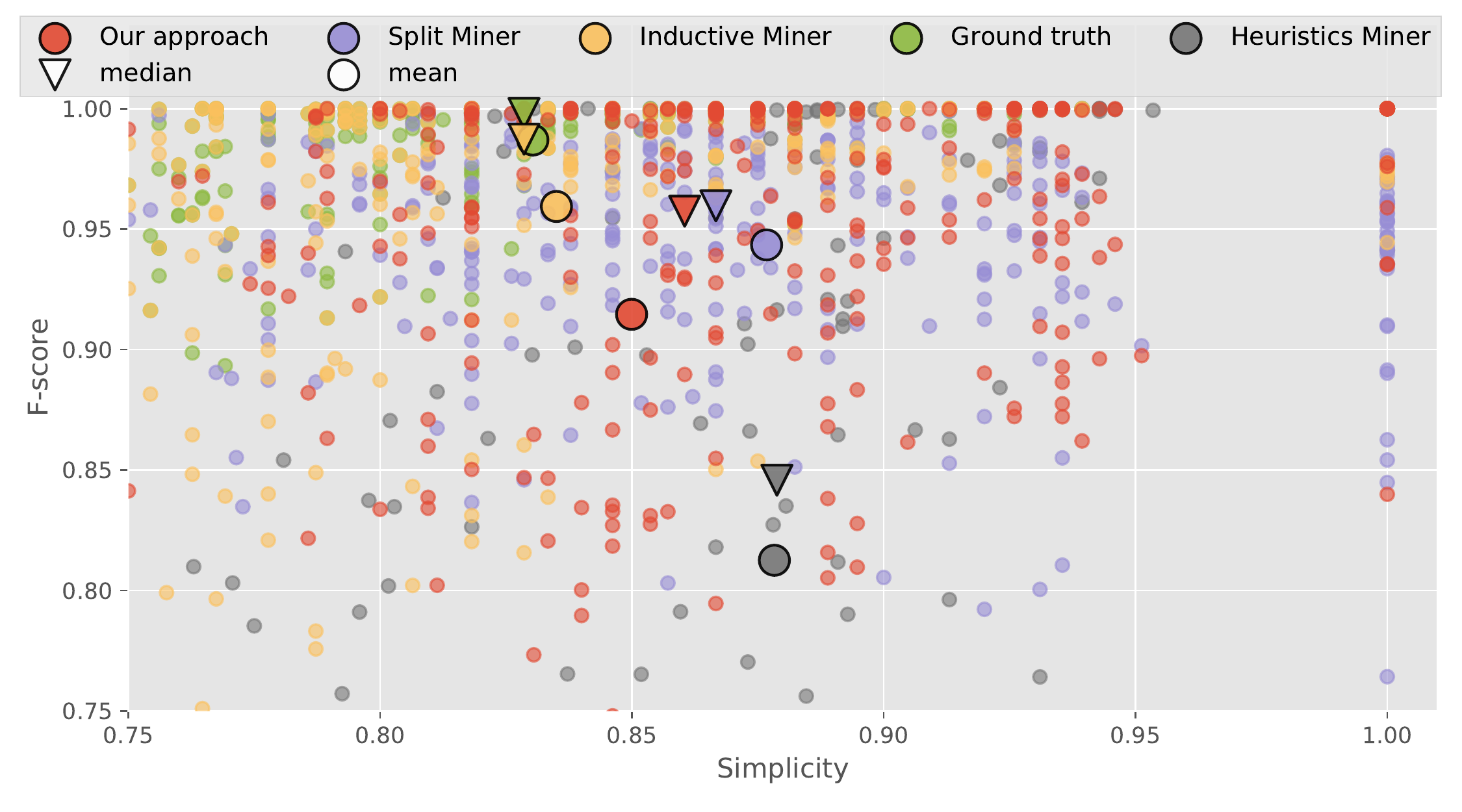}}
\caption{Conformance results on the synthetic test dataset with mean and median centroids.}
\label{fig:conf_synth_fs_s}
\end{figure}

\emph{(O2)} the scatter plot in Fig. \ref{fig:conf_synth_fs_s} shows the F-scores and simplicities of all produced models for the test dataset. Only for a subset of around 65\% did our approach find a valid process model, i.e. being at least easy sound and only their values are shown in Fig. \ref{fig:conf_synth_fs_s}. The medians and means for each method and the ground truth are shown as well indicating the average model qualities across the methods. Looking at the median centroids, IM clearly comes closest to the ground truth  (of block-structured models). Our approach and the SM discover simpler models with a small trade-off in F-score. HM lacks in F-score while achieving high simplicity. The large gap between the median and the mean for our approach shows that for a lot of samples, the conformance is very high and that for few samples the conformance is very low. Looking further into these, most of the lower scoring samples are not sound (only easy sound) and the processes are complex in structure: parallelism in the upper part of the process tree with multiple parallelism and loops in the lower parts, causing a large number of candidate places.

\begin{figure}[t]
\centerline{\includegraphics[width=1\linewidth]{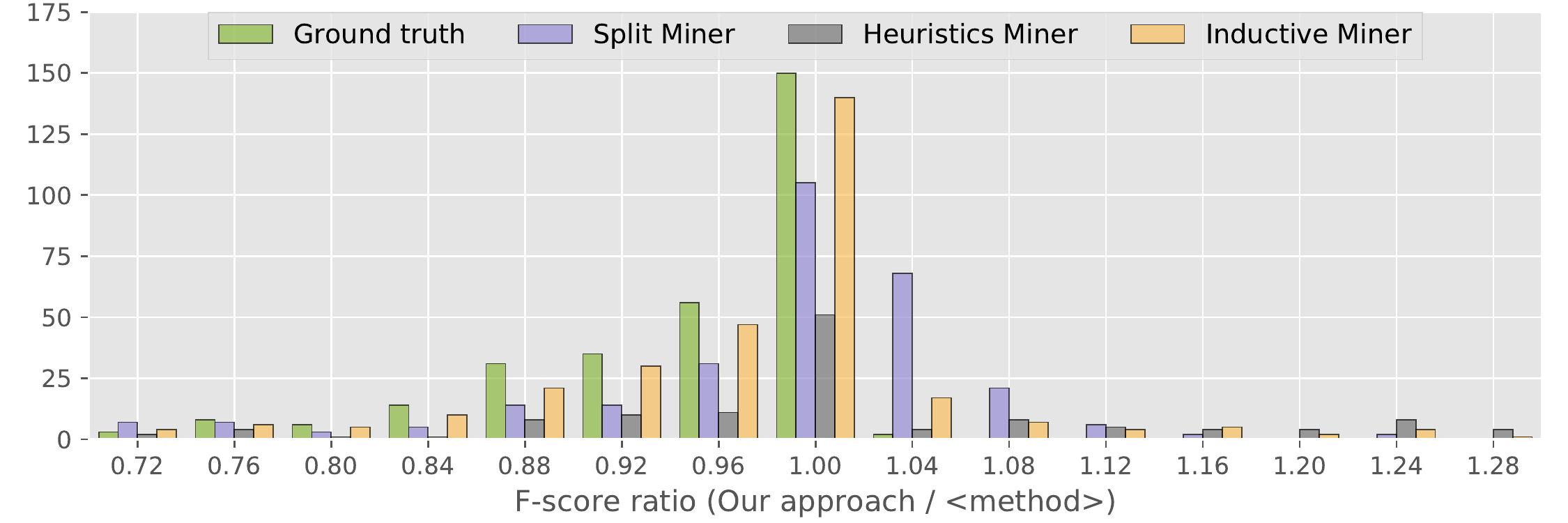}}
\caption{Histogram of F-score ratios comparing our approach to the ground truth and other methods.}
\label{fig:conf_synth_f_ratio_hist}
\end{figure}

To evaluate relative performance per log, we compute the F-score ratios of our approach over each other approach: a ratio $> 1$ means our approach performed better on a log. The histogram in Fig. \ref{fig:conf_synth_f_ratio_hist} shows that our approach outperforms SM on many logs while our approach struggles more significantly on a smaller subset of logs.


\begin{figure*}[t]
\centerline{\includegraphics[width=1\linewidth]{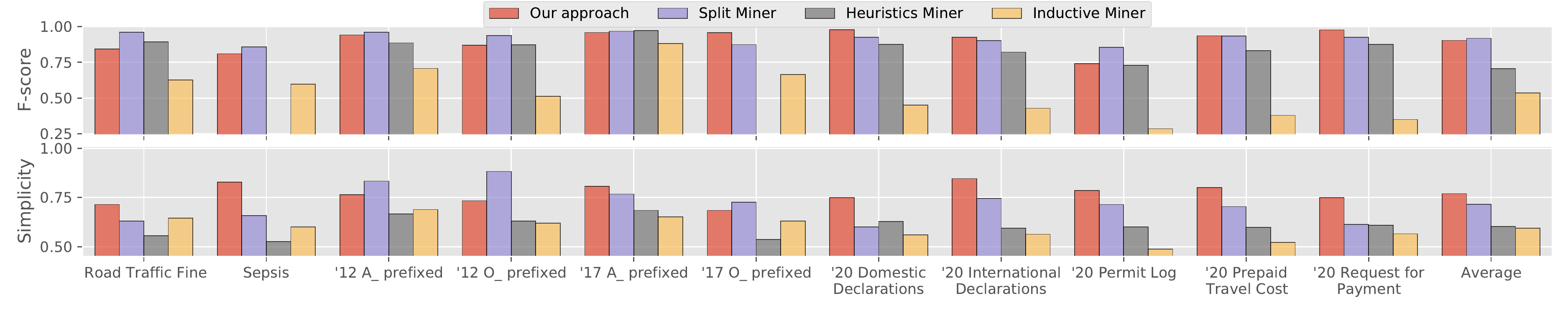}}
\caption{Conformance results on real-life datasets.}
\label{fig:conf_real_fs_s}
\end{figure*}

\emph{(O3)} Fig. \ref{fig:conf_real_fs_s} plots F-score and simplicity for each method and real-life dataset separately. The IM is known to problem to have low precision causing low F-scores. The SM generally achieves similar F-scores as the HM, but has higher simplicity. Our approach competes with the best scoring methods on all datasets in terms of F-scores and has the highest simplicity (except for BPIC'12); the average score plot in Fig. \ref{fig:conf_real_fs_s} reinforces this observation.

\begin{figure*}[t]
\centerline{\includegraphics[width=0.8\linewidth]{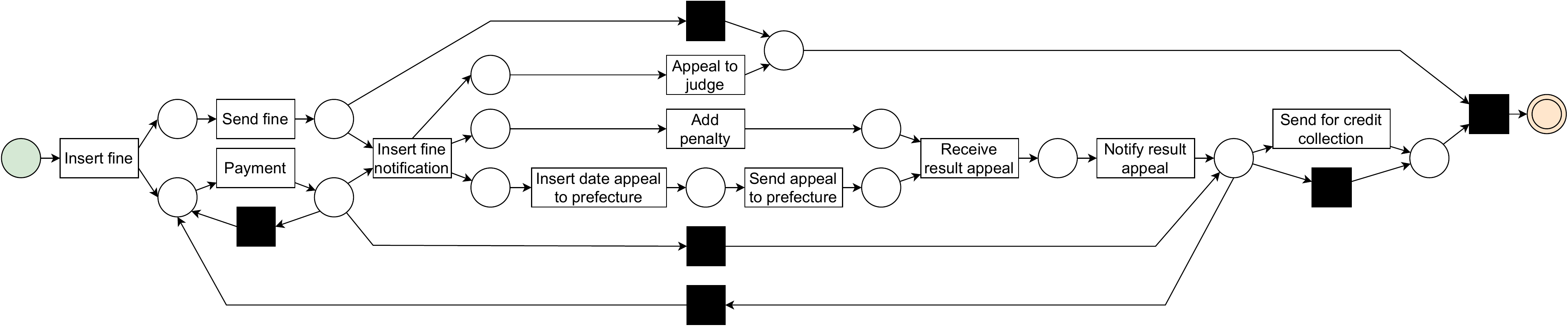}}
\caption{Discovered Petri net for the Road Traffic Fine dataset.}
\label{fig:result_rtf_gcn}
\end{figure*}

Fig. \ref{fig:result_rtf_gcn} shows the model discovered by our approach in the Road Traffic Fine dataset. The Petri net is not block structured but sound proving generalization beyond the block-structured training data. Furthermore, the model captures a complex synchronization of 2 parallel branches (repeated payments, sending fine) by an optional appeals procedure (with penalty in parallel) in an unstructured loop: payments can resume after conclusion of appeals.

Running times depend on several factors: creating graph $G$ from $L$ depends on the number of unique trace variants and candidate places (higher with more parallelism and loops). Place selection depends on the size of $G$ to compute embeddings on, the number of candidates to be chosen, and the beam width. Generally, the graph construction takes between 1 and 10 seconds with 250 and 2000 candidate places respectively and the candidate selection takes on average 1 second multiplied by the beam width. When $d$ has a hard time selecting places that keep the S-coverability, this can be substantially slower.

\section{Conclusion and Future Work}\label{sec:conclusion}
Our approach proved to perform well on the task it was trained on and generalizes beyond that exceeding the state-of-the-art methods on real-life datasets. This research laid down the foundation of introducing machine learning to APD and provides a framework in which various components can be optimized and replaced. Its design being based on a human modeler, which is still superior to automated methods, could enable new lines of APD research.

A limiting factor for this method to be scalable to large processes is the choice of the initial features, being one-hot encodings, allowing only a predefined maximum number of distinct activities. Future research is needed on a different label encoder that is scalable to the cardinality to counter this limitation.

Robustness has been a focus in automated process discovery since a wide variety of processes exist in terms of structure and algorithmic approaches often tackle only a specific kind of process. This has been a problem for our method as well where no sound model was discovered at all in some cases, which could be caused by the lack of generalization from the dataset it has been trained on, having a very specific representational bias. In our experiments, we reduced the number of traces to reduce the size of $G$ until a sound model was discovered. Although this is justified by the fact that human modelers also only look at the top most frequent/important traces and discover a model based on that, it is not ideal since possibly valuable information is lost. A first improvement here would be to increase the variety in the training data, but a proper fallback method should be in place to ensure robustness.


\bibliographystyle{elsarticle-num} 
\bibliography{apd-ml}

\begin{appendices}
\section{Reducing the search space}\label{app:search_space_reduction}
We can limit the set $P$ of candidate places to contain only those places which are compatible with the behavioral information in log $L$. For this, we use the $\alpha$-relations.

The \emph{$\alpha$-relations} over $A$~\cite{DBLP:journals/tkde/AalstWM04} serve as basis for many behavioral abstractions of an event log $L$ over $A$. Let $a,b \in A$. \emph{Directly-follows}: $a >_L b \iff \langle \ldots,a,b,\ldots\rangle \in L$; \emph{$k$-eventually follows}: $a >_L^k b \iff \langle \ldots,x_0,x_1,\ldots,x_k,\ldots\rangle \in L, x_0 = a, x_k = b$; \emph{causal relation}: $a \rightarrow_L b \iff a >_L b \wedge b \ngeq_L a$; \emph{conflict}: $a \#_L b \iff a \ngeq_L b \wedge b \ngeq_L a$; \emph{parallel}: $a \|_L b \iff a >_L b \wedge b >_L a$.

The $\alpha$-relations of $L$ over-approximate the behavior in $L$; behavior not described by the $\alpha$-relations is not in $L$ and $N$ does not have to describe it. Thus, we limited the candidate places $P$ to those places justified by the $\alpha$-relations, as follows.

For every two activities $a$ and $b$ where the eventually follows relations $a >^k_L b$ holds for a specified $k$, the one-to-one place $(\{a\},\{b\})$ is added. Note that all such places have a single incoming and outgoing transition, i.e., we define the set $P^{(1-1)}$ for a given maximum $K$ with
\begin{equation}\label{eq:1-1_places}
    (\{a\},\{b\}) \in P^{(1-1)} \textit{ iff } a,b \in A \wedge \exists 1 \leq k \leq K: a >_L^k b
\end{equation}


A one-to-many place $(\{a\},T_{out})$ with one incoming transition $\{a\}$ and many outgoing transitions $T_{out} = \{b_1,\ldots,b_k\}$ is constructed by combining one-to-one places $(\{a\},\{b_1\}),\ldots,(\{a\},\{b_k\})$ with the same incoming transition $\{a\}$, if and only if no two transitions in $b_i,b_j\in T_{out}$ have a parallel relation $b_i \|_L b_j$ as denoted in Equation \ref{eq:1-n_places}.
\begin{equation}\label{eq:1-n_places}
\begin{aligned}
    (\{a\},T_{out}) \in P^{(1-n)} \textit{ iff } T_{out} \subseteq \{ b \mid (\{a\},\{b\}) \in P^{(1-1)} \}\\
    \wedge |T_{out}| > 1
    \wedge \nexists_{b_1,b_2 \in T_{out}} b_1 \|_L b_2
\end{aligned}
\end{equation}
Many-to-one places $(T_{in},\{b\}) \in P^{(n-1)}$ are defined correspondingly.

Lastly, the many-to-many places $P^{(n-n)}$ are constructed by combining many-to-one and one-to-many places in a similar fashion:
\begin{equation}\label{eq:n-m_places}
\begin{aligned}
    (T_{in},T_{out}) \in P^{(n-n)} \textit{ iff } |T_{in}| > 1 \wedge |T_{out}| > 1 \wedge\\
    \left( (T_{in} \subseteq \{ a \mid (\{a\},T_{out}) \in P^{(1-n)} \} \wedge \nexists_{a_1,a_2 \in T_{in}} a_1 \|_L a_2)\right. \\
    \vee \left. (T_{out} \subseteq \{ b \mid (T_{in},\{b\}) \in P^{(n-1)} \} \wedge \nexists_{b_1,b_2 \in T_{out}} b_1 \|_L b_2) \right)
\end{aligned}
\end{equation}



\end{appendices}

\end{document}